# A Safety-Aware Role-Orchestrated Multi-Agent LLM Framework for Behavioral Health Communication Simulation

Ha Na Cho*
*Department of Informatics,
University of California, Irvine, USA
Chohn1@uci.edu

*Abstract*—Single-agent large language model (LLM) systems struggle to simultaneously support diverse conversational functions and maintain safety in behavioral health communication. We propose a safety-aware, role-orchestrated multi-agent LLM framework designed to simulate supportive behavioral health dialogue through coordinated, role-differentiated agents. Conversational responsibilities are decomposed across specialized agents, including empathy-focused, action-oriented, and supervisory roles, while a prompt-based controller dynamically activates relevant agents and enforces continuous safety auditing. Using semi-structured interview transcripts from the DAIC-WOZ corpus, we evaluate the framework with scalable proxy metrics capturing structural quality, functional diversity, and computational characteristics. Results illustrate clear role differentiation, coherent inter-agent coordination, and predictable trade-offs between modular orchestration, safety oversight, and response latency when compared to a single-agent baseline. This work emphasizes system design, interpretability, and safety, positioning the framework as a simulation and analysis tool for behavioral health informatics and decision-support research rather than a clinical intervention.

## I. INTRODUCTION

Recent advances in large language models (LLMs) have expanded the capabilities of conversational AI, enabling systems capable of empathetic dialogue. In digital mental health and behavioral health settings, such systems have been explored for applications including supportive communication, engagement facilitation, and decision-support simulation [1]. More recent work has extended these efforts by leveraging multi-agent LLM architectures for structured psychiatric interviews and assessment workflows, demonstrating the potential of role-specialized agent collaboration in mental health contexts [2]. However, most existing LLM-based dialogue systems adopt a single-agent architecture, which limits the capacity to represent the diverse and interdependent conversational functions involved in real-world counseling interactions. Supportive behavioral health communication is inherently multi-faceted, involving functions such as emotional validation, motivational encouragement, cognitive restructuring, and action planning, which are difficult to model transparently and reliably within a monolithic agent.

Prior studies have investigated various aspects of multi-agent LLM systems, including communication efficiency [3], [4], and systematic evaluation of coordination strategies across different agent interaction paradigms [5]. Other work has examined structured clinical reasoning and decision-support pipelines [6], [7], as well as multi-disciplinary collaboration through role-playing agents in healthcare settings [8], [9]. While these works demonstrate the benefits of agent-based decomposition, many frameworks emphasize task performance, protocol adherence, or internal reasoning chains rather than dynamic, role-coordinated, multi-turn dialogue.

Alongside advances in coordination and reasoning, recent research has increasingly highlighted the importance of safety and ethical safeguards in mental health-related dialogue systems. Prior studies have documented risks such as hallucination, inappropriate tone, and lack of transparency in LLM-based health applications [10]. More recent multi-agent designs, such as safety-focused agent architectures for mental health interaction, explicitly incorporate risk-aware monitoring mechanisms [11]. Nevertheless, continuous, real-time safety auditing remains insufficiently integrated into role-orchestrated, multi-agent conversational systems that operate over multiple dialogue turns.

To address these limitations, we propose a role-orchestrated multi-agent LLM framework designed to simulate supportive behavioral health communication. The framework assigns distinct conversational responsibilities to specialized agents and coordinates their activation through a controller that enforces safety auditing at every interaction. This work makes three contributions: (1) a modular multi-agent architecture that decomposes supportive communication into interpretable, role-specific components, (2) a safety-aware orchestration mechanism with persistent supervisory oversight, (3) a system-level evaluation using scalable proxy metrics to analyze response quality, computational efficiency, and design trade-offs. This study is positioned as a healthcare modeling and simulation contribution rather than a clinical intervention or therapeutic evaluation.

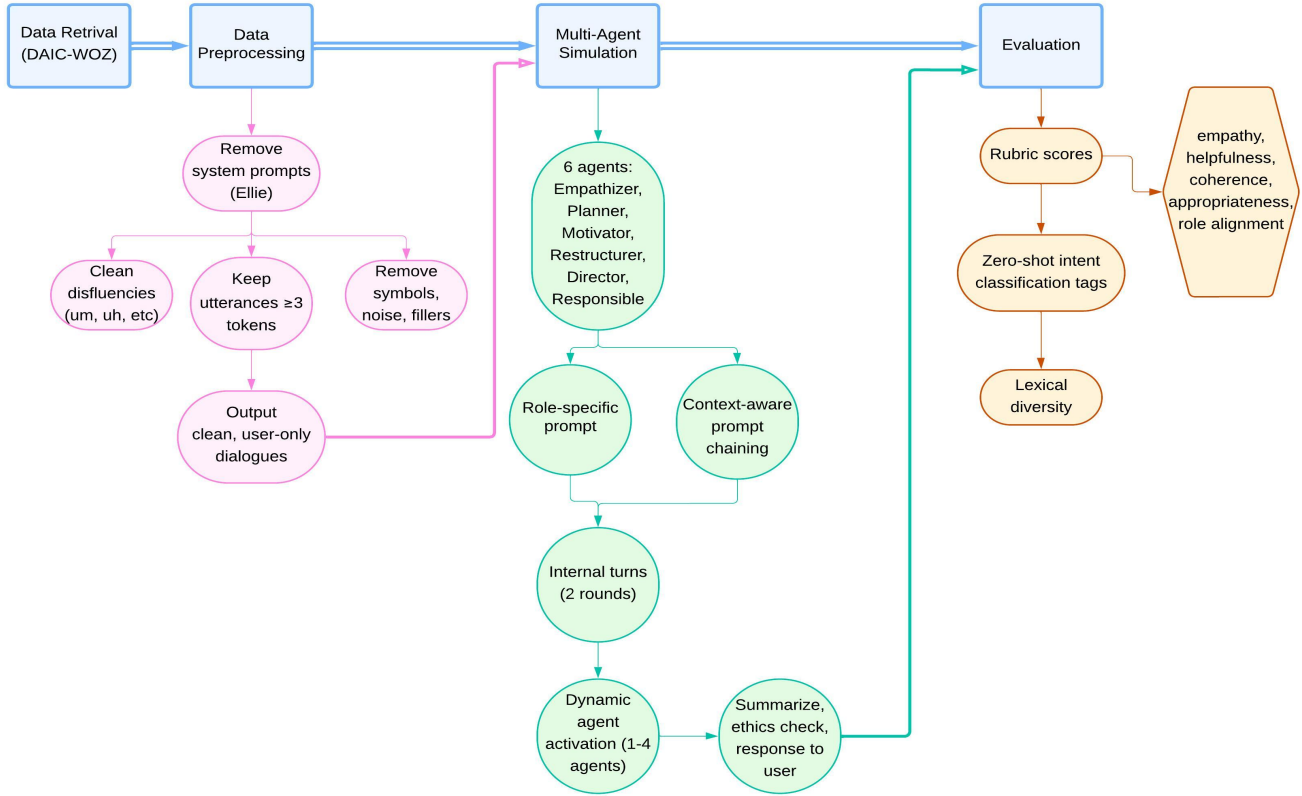

Fig. 1. Overview of the proposed role-orchestrated multi-agent simulation framework. Interview transcripts from the DAIC-WOZ dataset are preprocessed and provided to a prompt-based multi-agent system composed of six role-specialized agents. Role-specific prompts, context-aware prompt chaining, and dynamic agent activation govern multi-turn dialogue simulation. Generated responses are evaluated using rubric-based scoring, intent classification, and linguistic analysis.

## II. METHODS

### A. Dataset

We evaluated the proposed framework using the Distress Analysis Interview Corpus Wizard-of-Oz (DAIC-WOZ) dataset [12], [13]. DAIC-WOZ consists of semi-structured interviews conducted between participants and a virtual interviewer operated by a human through a Wizard-of-Oz protocol. The dataset includes synchronized audio, video, and text transcripts from 189 sessions. In this study, we used the transcript data only, focusing on participant utterances as inputs to the simulation framework. Seven participants were randomly selected to support manageable, qualitative system-level analysis. The selected sessions span a range of PHQ-8 severity scores, allowing examination of diverse linguistic and affective expressions. This subset was chosen to characterize system behavior and design trade-offs.

### B. Data Preprocessing

All transcripts were preprocessed to ensure consistency and compatibility with LLM inputs. Only participant utterances were retained; system responses from the virtual interviewer were excluded. Text cleaning was applied using regular expressions to remove non-lexical artifacts (e.g., laughter markers), special characters, and irregular spacing. Common disfluencies (e.g., "um," "uh") were removed to reduce noise while preserving syntactic structure. To focus on semantically meaningful inputs, utterances with fewer than three tokens after preprocessing were excluded. Short acknowledgments such as "okay" or "yeah" primarily serve turn-taking functions and provide limited contextual signal [14], [15]. These preprocessing steps were applied uniformly across all sessions.

### C. Role-Orchestrated Multi-Agent Simulation Framework

We propose a multi-turn, role-orchestrated multi-agent LLM framework for simulating supportive behavioral health communication. An overview of the system architecture is shown in Fig. 1. The system comprises six specialized agents: Empathizer, Motivator, Planner, Cognitive Restructurer, Director, and Responsible Agent. Each agent is instantiated with a role-specific prompt describing its conversational responsibility.

The Director is responsible for synthesizing agent-level outputs into a single user-facing response, while the Responsible Agent performs continuous safety auditing. Both supervisory roles are active at every dialogue turn to ensure structural consistency and emotional appropriateness. Agent activation is coordinated by a prompt-based controller that dynamically selects a subset of content-producing agents based on the current user utterance and recent agent outputs. Rather than relying on an explicit graph structure, coordination logic is encoded through predefined transition rules within the controller prompt. For example, expressions of emotional distress preferentially activate the Empathizer, while action-oriented content may trigger the Planner and Motivator. Safety auditing by the Responsible Agent is applied persistently across turns. Each user utterance is contextualized using a limited window of up to three preceding utterances and recent agent responses. Context windows are selectively filtered by role to maintain relevance and reduce redundancy. Agent outputs are recorded in a shared memory that supports inter-agent context propagation across turns. After agent-level reasoning, the Director synthesizes a single response, which is reviewed by the Responsible Agent before being returned as the system output. This operational flow for agent coordination and prompt generation at each turn is summarized in Table 1. All simulations were executed using GPT-3.5-turbo and GPT-4-turbo on an Apple M2 system (macOS 24.1, 11-core CPU, 18 GB RAM).

TABLE I. OPERATIONAL FLOW OF THE ROLE-ORCHESTRATED MULTI-AGENT SIMULATION PER USER TURN

| Step | Component | Description |
|---|---|---|
| 1 | User Input & Memory Update | For each user utterance, initialize or update the shared memory with the current input and recent dialogue context |
| 2 | Controller | For each dialogue turn (up to two iterations per user utterance), select active role agents based on prompt-encoded transition conditions derived from the user input and prior agent outputs |
| 3 | Role Agents | For each activated agent, construct a role-filtered dynamic context window and generate a role-specific response |
| 4 | Shared Memory | Store agent outputs to support inter-agent context propagation across turns |
| 5 | Director | Synthesize agent-level outputs into a single coherent user-facing response |
| 6 | Responsible Agent | Audit the synthesized response for emotional safety and ethical appropriateness |
| 7 | System Output | Return the audited response to the user and export agent activity logs for analysis |

### D. Evaluation Framework

We evaluated the system using a scalable, proxy-based evaluation framework designed for system-level analysis rather than clinical validation (Fig. 2). First, agent responses were assessed using a rubric-based scoring procedure implemented with GPT-4-turbo across five dimensions: empathy, helpfulness, coherence, appropriateness, and role alignment. Each dimension was scored on a 5-point Likert scale under fixed sampling settings to support reproducibility. The rubric dimensions were informed by established communication frameworks such as CARE and motivational interviewing [16], [17].

Second, therapeutic intent was analyzed using a zero-shot classification pipeline with GPT-3.5-turbo. Each response was categorized into one of twelve intent types (e.g., validation, encouragement, reflection, psychoeducation, coping suggestion, cognitive reframing, reassurance, empowerment, active listening, goal orientation, generic, and inappropriate), providing insight into role consistency and functional diversity.

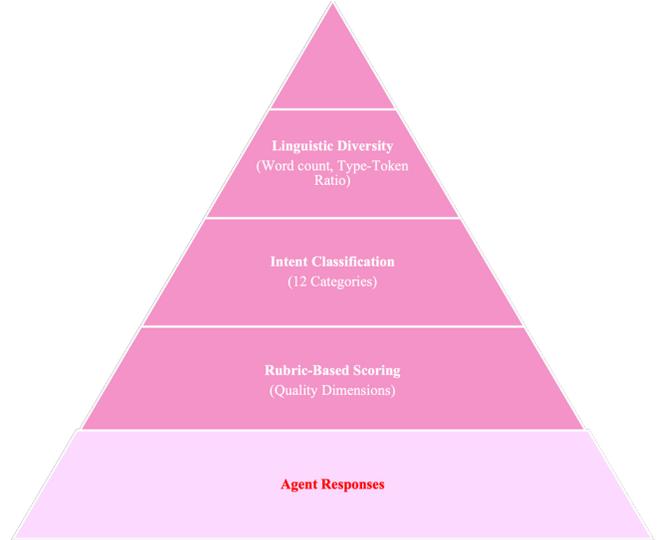

Fig. 2. Proxy-based evaluation framework for system-level analysis. General agent responses are evaluated across multiple abstraction levels, including rubric-based quality assessment, intent classification, and linguistic diversity metrics, to characterize system behavior without clinical outcome assessment.

Third, linguistic diversity was assessed using word count and type-token ratio (TTR) metrics to examine whether different agent roles contributed non-redundant language patterns. This evaluation framework prioritizes consistency, interpretability, and computational efficiency, and is intended to characterize system behavior and design trade-offs.

## III. RESULTS

### A. Descriptive Characteristics of Data

We conducted a descriptive analysis of seven participants (IDs: 300, 319, 361, 396, 446, 480, 492) sampled from the DAIC-WOZ corpus to contextualize the linguistic and structural properties of the dialogue data used for system-level evaluation. Across participants, the number of user utterances ranged from 75 to 221, with total session lengths (including interviewer turns) ranging from 118 to 295 utterances. Participant-generated speech accounted for approximately 60%-75% of total dialogue turns, reflecting variability in interaction dynamics and responsiveness.

Average utterance length varied substantially across participants, indicating heterogeneous expressive styles. Participant 361 exhibited the longest mean utterance length (17.36 tokens), whereas participant 300 showed the shortest (4.05 tokens). This diversity in utterance length and participation patterns provided a heterogeneous input space for evaluating agent activation, coordination behavior, and response generation across different conversational contexts.

### B. Agent Activation Patterns and Inter-Agent Coordination

To examine how the role-orchestrated framework behaved under context-conditioned control, we analyzed agent activation frequencies across all user utterances. As shown in Fig. 3, the Director and Responsible Agent were activated most frequently (N = 1370 each), reflecting their fixed supervisory roles at every dialogue turn. In contrast, content-generating agents, including the Empathizer, Motivator, and Planner, were activated selectively based on user input and evolving conversational context.

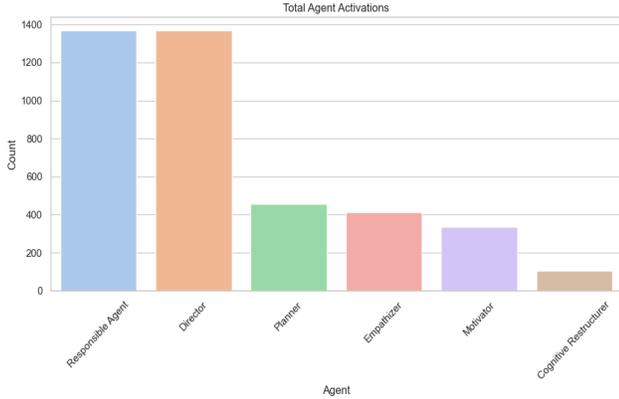

Fig. 3. Frequency of agent activations across all dialogue turns. Supervisory agents are activated at every turn by design, while content-generating agents are selectively invoked based on context-conditioned controller decisions.

The Cognitive Restructurer was invoked less frequently than other content agents, indicating that explicit cognitive reframing was not prioritized for the majority of user utterances in this sample. This distribution reflects both the controller's prompt-encoded transition rules and the nature of the underlying dialogue data, rather than fixed role scheduling. To further assess coordination behavior, we analyzed agent-to-agent transition patterns. As shown in Fig. 4, outputs from content-generating agents most frequently transitioned into Director responses, particularly from the Responsible Agent (663 transitions), demonstrating structured inter-agent chaining within the supervisory design. Transitions from the Empathizer (341) and Motivator (196) were also observed, indicating their contributions to synthesized responses. However, the Cognitive Restructurer exhibited limited downstream transitions, showing reframing-focused outputs were less frequently followed by subsequent agent activity. These patterns indicate coherent, role-consistent coordination dynamics, as opposed to uniform agent coordination.

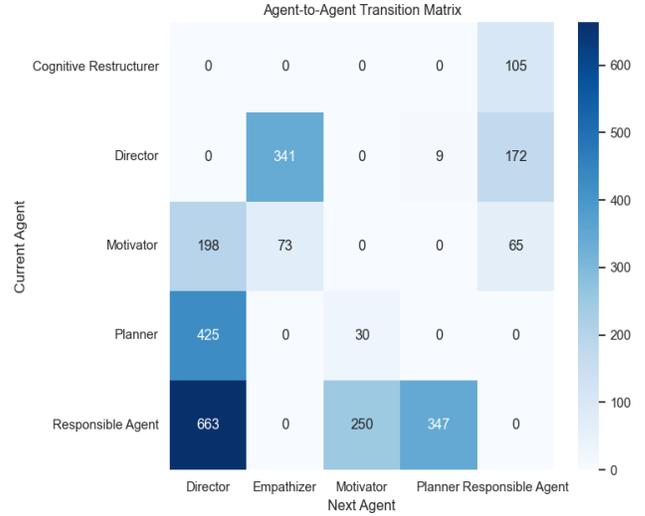

Fig. 4. Inter-agent transition patterns illustrating coordination behavior. Cell values indicate how often outputs from one agent are followed by responses from another.

## C. Computational Efficiency and Resource Use

We next examined computational efficiency by analyzing generation latency and token usage for each agent role. Distinct efficiency profiles emerged that reflected role responsibilities within the architecture. The Director exhibited the highest average latency (approximately 3.5 seconds), consistent with its synthesis fucntion and frequent invocation at each dialogue turn. Despite higher latency, Director responses remained relatively concise, indicating a focus on coordination and summarization rather than verbose output.

In contrast, affect-oriented agents such as the Empathizer and Motivator generated longer responses with higher token counts but lower average latency, reflecting expressive and supportive language generation. Procedural agents, including the Planner and Cognitive Restructurer, demonstrated both low latency and low token usage, suitability for high-frequency or resource-constrained settings.

TABLE II. OPERATIONAL FLOW OF THE ROLE-ORCHESTRATED MULTI-AGENT SIMULATION PER USER TURN

| Agent Role | Activation Pattern | Empathy | Helpfulness | Coherence | Appropriateness | Role Alignment | Lexical Diversity (TTR) |
|---|---|---|---|---|---|---|---|
| Empathizer | Selective | 4.80 | 3.80 | 5.00 | 5.00 | 5.00 | 0.13 |
| Motivator | Selective | 4.00 | 4.00 | 5.00 | 4.83 | 5.00 | 0.15 |
| Planner | Selective | 3.60 | 3.80 | 5.00 | 4.80 | 4.60 | 0.14 |
| Cognitive Restructurer | Rare | 4.00 | 4.00 | 5.00 | 5.00 | 5.00 | 0.24 |
| Director | Persistent | 4.00 | 4.11 | 5.00 | 5.00 | 5.00 | 0.07 |
| Responsible | Persistent | 3.86 | 4.00 | 5.00 | 4.93 | 5.00 | 0.08 |

*D. Linguistic Diversity Across Agent Roles*

Linguistic diversity was assessed using surface-level text statistics, TTR and word count metrics, to examine whether different agent roles contributed distinct language patterns. The Cognitive Restructurer exhibited the highest TTR (0.24), indicating greater lexical variability, consistent with its reframing-oriented function. In contrast, supervisory roles showed the lowest diversity (Director: 0.07; Responsible Agent: 0.08), while Empathizer (0.13), Motivator (0.15), and Planner (0.14) demonstrated moderate lexical variation.

*E. Structured Qualitative and Intent-Level Evaluation*

To further characterize response quality and functional diversity, we conducted rubric-based scoring and intent classification on a stratified subsample of 50 generated responses. Using a proxy-based rubric, responses were evaluated across five dimensions: empathy, helpfulness, coherence, appropriateness, and role alignment (Table 2). Across all agent roles, coherence and role alignment scores were consistently high (mean = 5.00), indicating structurally sound and role-consistent outputs under the evaluation framework.

Affective dimensions varied by role. The Empathizer achieved the highest mean empathy score (4.80), while more procedural roles such as the Planner and Responsible Agent showed lower empathy scores (3.60 and 3.86, respectively). The Director demonstrated balanced performance across all dimensions, reflecting its synthesis-focused function.

In parallel, each response was classified into one of twelve predefined therapeutic intent categories. The most frequent intents were psychoeducation (34.0%), empowerment (20.0%), and encouragement (14.0%), reflecting an emphasis on informational and motivational strategies. Less frequent but functionally important intents, including validation (8.0%), reflection (4.0%), coping suggestion (4.0%), and cognitive reframing (6.0%), appeared in responses associated with emotionally complex user inputs. All twelve intent categories were observed at least once, indicating full coverage of the predefined intent space and alignment between role conditioning and functional output.

## IV. DISCUSSION

This study provides system-level evidence that role-orchestrated multi-agent prompting can serve as a viable design paradigm for safety-sensitive dialogue modeling, beyond conversational single-agent or loosely coordinated multi-agent approaches [18], [19]. Our findings highlight orchestration behavior itself as a first-class design variable. The observed coordination patterns demonstrate that how agents are activated, sequenced, and supervised meaningfully shapes dialogue structure, functional diversity, and safety characteristics.

A key insight from the analysis is the emergence of clear functional differentiation across agent roles. Affect-oriented agents primarily contributed empathetic language, while procedural agents emphasized planning and synthesis. This alignment between prompt specification and generated behavior suggests that explicit role conditioning can reliably constrain conversational functions, consistent with prior work on role-based prompting and controllable generation [20], [21]. At the same time, affective competence remained largely concentrated within a single agent, indicating a compartmentalization of emotional reasoning rather than its integration across the system. This design choice supports interpretability and control but also reveals a limitation: emotional sensitivity is not yet a shared property of the dialogue as a whole. Future orchestration strategies may benefit from allowing affective signals to influence multiple roles dynamically rather than remaining isolated.

Intent-level analysis further contextualizes these patterns. The dominance of informational and motivational intents reflects conservative controller behavior favoring solution-oriented responses, also observed in efficiency-driven frameworks [3], [5]. While such behavior may be appropriate for coaching-style interactions, it risks underrepresenting reflective or emotionally supportive strategies when deeper affective engagement is warranted. Importantly, the presence of all predefined intent categories indicates that the system is capable of expressing a broad functional space, even if its current scheduling biases limit their relative frequency. This distinguishes the proposed framework from prior systems that restrict conversational scope through fixed agent pipelines [22]. This observation motivates adaptive role weighting or context-aware activation policies as promising extensions.

Prompt design emerged as a central determinant of inter-agent coordination. Early configurations that relied on generic response instructions produced fragmented dialogue with minimal cross-agent dependency. Explicitly instructing agents to reference prior peer outputs, however, enabled structured chaining and improved coherence. Nevertheless, some roles, particularly those associated with cognitive reframing, exhibited limited downstream influence, suggesting that not all conversational functions proposed equally under static orchestration rules. This finding contrasts with the system that rely on iterative consensus or debate mechanisms [8], which show the need for more flexible controllers that can account for temporal context and evolving conversational trajectories.

Safety considerations are addressed at the architectural level rather than through post-hoc filtering. The persistent inclusion of a dedicated Responsible Agent and the conditional activation of empathy-focused roles constitute structural safeguards that monitor tone, emotional risk, and ethical appropriateness throughout the dialogue. Unlike approaches that treat safety as an external constraint, this design embeds oversight directly into the generation process, making safety behavior observable and auditable. While this study does not evaluate real-world user impact, the framework provides a controlled environment for examining safety mechanisms in multi-agent dialogue systems.

This study has several limitations. The framework operates entirely in a simulated setting and relies on proxy-based evaluation rather than human judgment. Dialogue coordination is sequential rather than conversational among agents, limiting the diversity of perspectives that could emerge from richer inter-agent interaction. Context filtering trades efficiency for sensitivity and may omit short utterances with affective significance. These limitations are not shortcomings of implementation but reflections of deliberate design choices made to prioritize interpretability, safety, and reproducibility.

## V. Conclusion

This work presents a role-orchestrated multi-agent LLM framework as a modeling approach for analyzing supportive behavioral health communication under explicit safety and interpretability constraints. By structuring dialogue generation around specialized roles and supervisory oversight, the framework provides a controllable and inspectable alternative to monolithic conversational agents in sensitive application contexts. The design prioritizes transparency in how conversational functions are allocated and coordinated, enabling systematic examination of dialogue behaviors and system-level trade-offs. This framework serves as a simulation platform for studying agent coordination and functional differentiation in language-based systems. Future work will explore adaptive orchestration strategies, richer inter-agent interaction patterns, and human-in-the-loop evaluation to further refine system behavior and applicability.